\documentclass[sigconf]{acmart}

\AtBeginDocument{%
  \providecommand\BibTeX{{%
    \normalfont B\kern-0.5em{\scshape i\kern-0.25em b}\kern-0.8em\TeX}}}

\setcopyright{acmcopyright}
\copyrightyear{2020}
\acmYear{2020}
\acmDOI{}

\acmConference[WSDM '21]{ACM International Conference on Web Search and Data Mining}{March 08-12, 2021}{Jerusalem, Israel}
\acmPrice{15.00}
\acmISBN{978-1-4503-XXXX-X/18/06}

\usepackage{amsmath,amssymb,amsthm,amsfonts,mathtools,color,nicefrac,multirow,bbm,rotating,booktabs,subfigure}
\usepackage{enumitem}
\usepackage[linesnumbered, ruled]{algorithm2e}

\newcommand{\R}{\mathbb{R}}

\begin{document}

\title{Explainable Multivariate Time Series Classification: A Deep Neural Network Which Learns To Attend To Important Variables As Well As Informative Time Intervals}
\renewcommand{\shorttitle}{Explainable Multivariate Time Series Classification: LAXCAT}

\author{Tsung-Yu Hsieh}
\affiliation{%
  \institution{The Pennsylvania State University}
  \country{University Park, PA, USA}}
\email{tuh45@psu.edu}

\author{Suhang Wang}
\affiliation{%
  \institution{The Pennsylvania State University}
  \country{University Park, PA, USA}}
\email{szw494@psu.edu}

\author{Yiwei Sun}
\affiliation{%
  \institution{The Pennsylvania State University}
  \country{University Park, PA, USA}
}
\email{yus162@psu.edu}

\author{Vasant Honavar}
\affiliation{%
  \institution{The Pennsylvania State University}
  \country{University Park, PA, USA}}
\email{vhonavar@psu.edu}


\begin{abstract}
Time series data is prevalent in a wide variety of real-world applications and it calls for trustworthy and explainable models for people to understand and fully trust decisions made by AI solutions. We consider the problem of building explainable classifiers from multi-variate time series data.  A key criterion to understand such predictive models involves elucidating and quantifying the contribution of time varying input variables to the classification. Hence, we introduce a novel, modular, convolution-based feature extraction and attention mechanism that simultaneously identifies the variables as well as time intervals which determine the classifier output. We present results of extensive experiments with several benchmark data sets that show that the proposed method outperforms the state-of-the-art baseline methods on multi-variate time series classification task. The results of our case studies demonstrate that the variables and time intervals identified by the proposed method make sense relative to available domain knowledge.
\end{abstract}





\maketitle

\section{Introduction}

Recent advances in high throughput sensors and digital technologies for data storage and processing have resulted in the availability of complex multivariate time series (MTS) data, i.e., measurements from multiple sensors, in the simplest case, sampled at regularly spaced time points, that offer traces of complex behaviors as they unfold over time. There is much interest in effective methods for classification of MTS data \cite{bagnall2017great} across a broad range of application domains including finance~\cite{wu2013dynamic}, metereology~\cite{chakraborty2012fine}, graph mining~\cite{xuan2007modeling, sun2019megan}, audio representation learning~\cite{franceschi2019unsupervised, sun2020ontology}, healthcare~\cite{lipton2015learning, liang2020lmlfm, dai2020ginger}, human activity recognition~\cite{martinez2017human, tang2020joint, hsieh2019adaptive}, among others. The impressive success of deep neural networks on a broad range of applications~\cite{lecun2015deep} has spurred the development of several deep neural network models for MTS classification~\cite{fawaz2019deep}. For example, recurrent neural network and its variants LSTM and GRU are the state-of-the-art methods for modeling the complex temporal and variable relationships~\cite{hochreiter1997long,cho2014properties}. 
 
In high-stakes applications of machine learning, the ability to explain a machine learned predictive model is a prerequisite for establishing {\em trust} in the model's predictions, and for gaining scientific insights that enhance our understanding of the domain. MTS classification models are no exception: In healthcare applications, e.g., monitoring and detection of epileptic seizures, it is important for clinicians to understand how and why an MTS classifier classifies EEG signal as indicative of onset of seizure~\cite{shoeb2009application}. Similarly, in human activity classification, it is important to be able to explain why an MTS classifier detects activity that may be considered suspicious or abnormal~\cite{yin2008sensor}. Although there has been much recent work explaining black box predictive models and their predictions~\cite{mueller2019explanation,guidotti2018survey,khademi2020causal}, the existing methods are not directly applicable to MTS classifiers.

\begin{figure}
    \centering
    \subfigure[Normal example]{
    \includegraphics[width=0.43\textwidth]{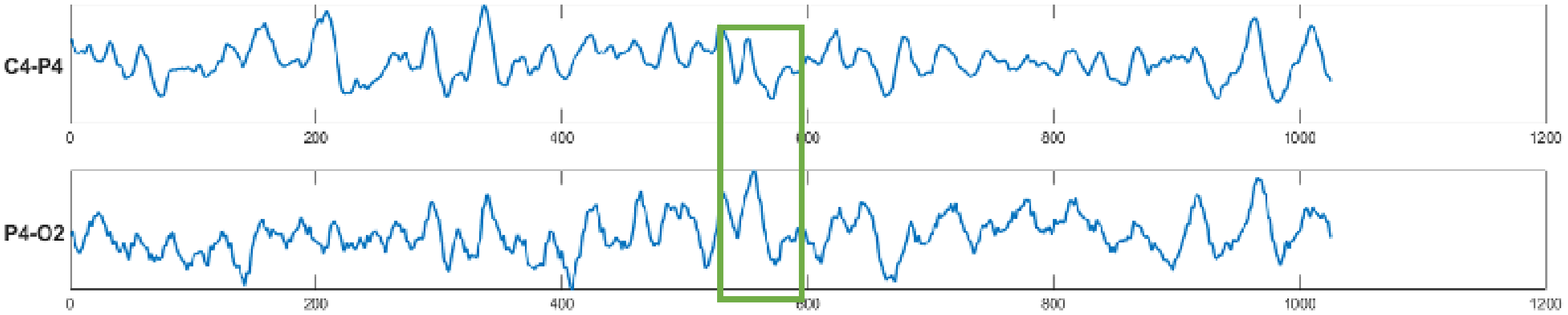}
        \label{fig:toy_example:healthy}
    }\\
    \subfigure[Seizure example]{
    \includegraphics[width=0.43\textwidth]{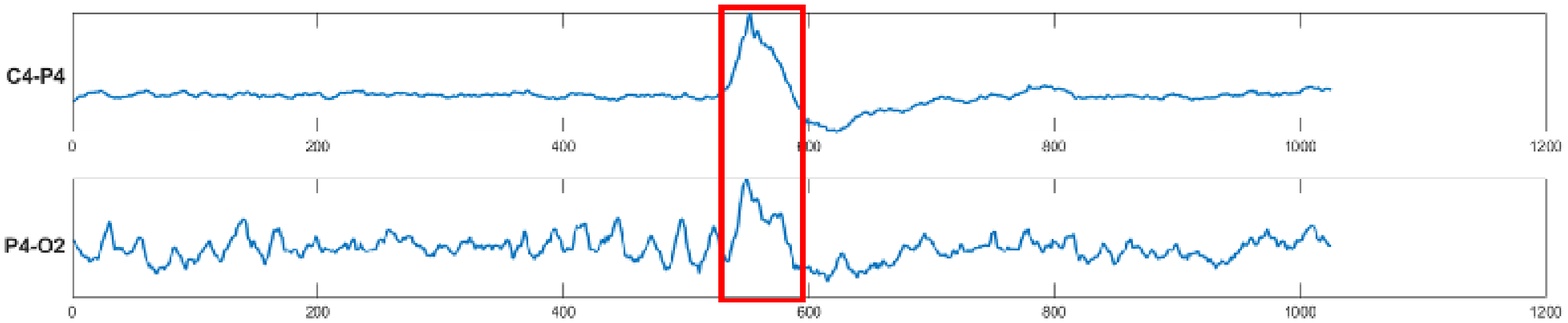}
        \label{fig:toy_example:seizure}
    }
    \vskip -2em
    \caption{Normal and seizure brain wave signal examples.}
    \label{fig:toy_example}
    \vspace{-10pt}
    \vskip -1em
\end{figure}

Developing explainable MTS data presents several unique challenges: Unlike in the case of classifiers trained on static data samples, MTS data encode {\em the patterns of variable progression over time}. For example, compare the brain wave signals (electroencephalogram or EEG recordings) from a healthy patient with those from a patient suffering from epileptic seizure as shown in Figure~\ref{fig:toy_example}~\cite{shoeb2009application,goldberger2000physiobank}. The two EEG recordings differ with respect to the temporal patterns in the signals~\cite{major2007seizures}. Because EEG measurements obtained at high temporal resolution suffer from low signal-to-noise ratio, the EEG recordings from healthy patients (see Figure~\ref{fig:toy_example:healthy}) display some of the spike-like signals that are similar to those indicative of seizure Figure~\ref{fig:toy_example:seizure}. However, the temporal pattern of EEG signals over a larger time window shows clear differences between healthy and seizure activity. Thus, undue attention to local, point-wise observations, without consideration of the entire temporal pattern of activity~\cite{li2019enhancing} would result in failure to correctly recognize abnormal EEG recordings that are indicative of seizure. In contrast, focusing on the temporal pattern of activity over the relevant time windows as shown in Figure~\ref{fig:toy_example}, would make it easy to distinguish the EEG recordings indicative of healthy brain activity from those that are indicative of seizure, and to explain how they differ from each other.  In the case of MTS data, each variable offers different amounts of information that is relevant to the classification task. Furthermore, different variables may provide discriminative information during different time intervals. Hence, we hypothesize that MTS classifiers that can simultaneously identify not only important variables but also the time intervals during which the variables facilitate effective discrimination between different classes can not only improve the accuracy of MTS classifiers, but also enhance their explainability.

Hence, we introduce a novel, modular, convolution-based feature extraction and attention mechanism that simultaneously (i) identifies informative variables and the time intervals during which they contain informative patterns for classification; and (ii) leverages the informative variables and time intervals to perform MTS classification. Specifically, we propose Locality Aware eXplainable Convolutional ATtention network (LAXCAT), a novel MTS classifier which consists of dedicated convolution-based feature extraction network and dual attention networks. The convolution feature extraction network extracts and encodes information from a local neighborhood around a time point. The dual attention networks help identify the informative variables and the time intervals in which each variable helps discriminate between classes. Working in concert, the convolution-based feature extraction network and the dual attention networks maximize predictive performance and the explainability of the MTS classifier. The major contributions of this work are as follows:
\begin{itemize}
    \item We consider the novel problem of simultaneously selecting informative variables and identifying the time intervals that define informative patterns of values for discrimination between the classes to optimize the accuracy and explainability of MTS classifiers;
    \item We describe a novel modular architecture consisting of a convolution-based feature extraction network and dual attention networks that offers an effective solution to this problem.
    \item We present results of extensive experiments with several benchmark data sets that the proposed method outperforms the state-of-the-art baseline methods for multi-variate time series classification. 
    \item We present results of case studies and demonstrate that the variables and time intervals identified by the proposed model are in line with the available domain knowledge.
\end{itemize}
The rest of the paper is organized as follows. Section 2 reviews related work; Section 3 introduces the problem definition; Section 4 describes LAXCAT, our proposed solution; Section 5 describes our experiments and case studies; Section 6 concludes with a brief summary and discussion of some directions for further research.

\section{Related Work}
\noindent \textbf{Multivariate Time Series Classification.} Multi-variate time series classification has received much attention in recent years. Such methods can be broadly grouped into two categories: distance-based methods~\cite{abanda2019review} and feature-based methods~\cite{fulcher2014highly}. Distance-based methods classify a given time series based on the label(s) of the time series in the training set that are most {\em similar} to it or closest to it where closeness is defined by some distance measure. Dynamic time warping (DTW)~\cite{berndt1994using} is perhaps the most common distance measure for assessing the similarity between time series. DTW, combined with the nearest neighbors (NN) classifier is a very strong baseline method for MTS classification~\cite{bagnall2017great}. Feature-based methods extract a collection of informative features from the time series data and encode the time series using a feature vector. The simplest such encoding involves representing the sampled time series values by a vector of numerical feature values. Other examples of time series features include various statistics such as sample mean and variance, energy value from the Fourier transform coefficients, power spectrum bands~\cite{bloomfield2004fourier}, wavelets~\cite{percival2000wavelet}, {\em shapelets}~\cite{ye2009time}, among others. Once time series data are encoded using finite dimensional feature vectors, the resulting data can be used to train a classifier using any standard supervised machine learning method~\cite{hastie2009elements}. The success of deep neural networks on a wide range of classification problems~\cite{lecun2015deep} has inspired much work on variants of deep neural networks for time series classification (see~\cite{fawaz2019deep} for a review). However, as noted earlier, the black box nature of deep neural networks makes them difficult to understand. Deep neural network models for MTS classification are no exceptions. 

\noindent \textbf{Explainable Models.} There has been much recent work on methods for explaining black box predictive models (reviewed in~\cite{mueller2019explanation,khademi2020causal}, typically, by attributing the responsibility for the model's predictions to different input features. Such post hoc model explanation techniques include methods for visualizing the effect of the model inputs on its outputs~\cite{yosinski2015understanding,zeiler2014visualizing}, methods for extracting simplified rules or feature interactions from black box models~\cite{frosst2017distilling,murdoch2018beyond}, methods that score features according to their importance in prediction~\cite{lundberg2017unified,ancona2019explaining,chen2018learning}, gradient based methods that assess how changes in inputs impact the model predictions \cite{shrikumar2016not,shrikumar2017learning,chen2018learning}, and methods for approximating local decision surfaces in the neighborhood of the input sample via localized regression \cite{ribeiro2016should,selvaraju2017grad,bhatt2020explainable}.  

An alternative to post hoc analysis is explainability by design, which includes in particular, methods that identify an informative subset of features to build parsimonious, and hence, easier to understand models. Such methods can be further categorized into {\em global} methods which discover a single, instance agnostic subset of relevant variables, and {\em local} methods which discover instance-specific subsets of relevant features. Yoon et al.~\cite{yoon2006feature} proposed a principal component analysis-based recursive variable elimination approach to identify informative subset of variables on an fMRI classification task. Han et al. ~\cite{han2013feature} use class separability to select optimal subset of variables in a MTS classification task. When the data set is heterogeneous, it may be hard to identify a single set of features that are relevant for classification over the entire data set ~\cite{yoon2018invase}. Such a setting calls for local methods that can identify instance-specific features. One such local method uses an attention mechanism ~\cite{bahdanau2015neural}. Choi et al.~\cite{choi2016retain} proposed RETAIN, an explainable predictive model based on a two-level neural attention mechanism which identifies significant clinical variables and influential visits to the hospital in the context of electronic health records classification. RAIM~\cite{xu2018raim} introduced a multi-channel attention mechanism guided by discrete clinical events to jointly analyze continuous monitoring data and discrete clinical events. Qin et al.~\cite{qin2017dual} proposed a dual-stage attention-based encoder-decoder RNN to select the time series variables that drive the model predictions. Guo et al.~\cite{guo2019exploring} explored the structure of LSTM networks to learn variable-wise hidden states to understand the role of each variable in the prediction.
A key limitation of the existing body of work on explaining black box neural network models for MTS classification is that they focus on either identifying a subset of relevant time series, or a subset of discrete time points. However, many practical applications of MTS classification, require identifying not only the relevant subset of the time series variables, but also the  time intervals during which the variables help discriminate between the classes.  

\begin{figure*}[t]
    \centering
    \includegraphics[width = 0.96\textwidth]{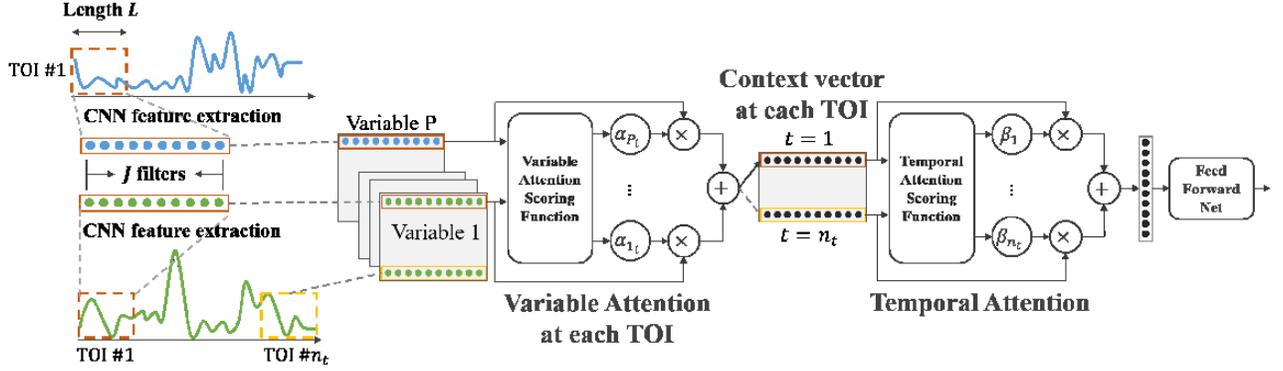}
    \caption{Proposed LAXCAT model framework. The framework is comprised of three major components: CNN feature extraction module and two attention modules. The CNN layer extracts informative features within each time interval of interest (TOI). The two attention modules work together to identify informative variables and key TOIs.}
    \vskip -1em
    \label{fig:FATCAT_flowchart}
\end{figure*}

\section{Problem Definition}

Let $\mathbf{X} = \lbrace \mathbf{x}^{(1)},\ldots,\mathbf{x}^{(T)} \rbrace$ be a multivariate time series sequence, where $\mathbf{x}^{(t)} \in \R^P$ denotes the $P$ dimensional observation at time point $t$. $\mathbf{x}^{(t)}_i \in \R$ means the value of the $i$-th variable sampled at time point $t$. We use $\mathcal{X} = \left\lbrace (\mathbf{X}_1,y_1),\ldots,(\mathbf{X}_N,y_N) \right\rbrace$ to denote a set of $N$ input sequences along with their true labels, where $\mathbf{X}_i$ is the $i$-th multivariate time series sequence and $y_i$ is its corresponding label.  Based on the context, $t$ can be used to index either a time point or a time interval.
In multivariate time series classification (MTSC), the goal is to predict the label $y$ of a MTS data $\mathbf{X}$. For example, given sequences of EEG recordings of a subject from multiple channels corresponding to different locations on the brain surface, the task is to predict whether it denotes healthy or seizure activity. As noted earlier each multivariate time series, not all the features equally inform the classification. In addition, for the important variables, only few key time intervals are typically important for discrimination between the different classes. Hence the problem of explainable MTSC is formally defined as follows
\vskip 0.5em
\noindent{}\textit{Given an MTS training data set $\mathcal{X} = \left\lbrace (\mathbf{X}_1,y_1),\ldots,(\mathbf{X}_N,y_N) \right\rbrace$, learn a function $f$ that can simultaneously predict the label of a MTS data and identify the informative variables as well as the time intervals over which their values inform the class label assigned to each multivariate time series. 
}

\section{The Proposed Framework - LAXCAT}
We proceed to describe Locality Aware eXplainable Convolutional ATtention network (LAXCAT). Figure~\ref{fig:FATCAT_flowchart} provides an overview of the LAXCAT architecture. LAXCAT consists of three components: (i) a convolutional module that extracts time-interval based features from the input multivariate time series sequence; (ii) variable attention module, which assigns weights to variables according to their importance in classification; and (iii) temporal attention module, which identifies the  time intervals over which the variables identified by the variable attention module inform the classifier output.
The LAXCAT architecture is designed to learn a representation of the MTS data that not only suffices for accurate prediction of the class label for each MTS data instance, but also helps explain the assigned class label in terms of the variables {\em and} the time intervals over which the values they assume inform the classification. We now proceed to describe each module of LAXCAT in detail.

\subsection{Feature Extraction via Convolutional Layer} \label{sec:feature_extraction}
The first step is to extract useful features from the input time series. The key idea of the feature extraction module is to incorporate temporal pattern of values assumed by a time series variable  as opposed to focusing only on point-wise observations. Given a multivariate time series input sequence $\mathbf{X} = \left\lbrace \mathbf{x}^{(1)}, \ldots, \mathbf{x}^{(T)} \right\rbrace$, with $\mathbf{x}^{(t)} \in \R^P$, where $T$ is the length of the sequence and $P$ is the number of covariates, we adopt convolutional layer to automatically extract features from the time series.  
Specifically, a 1-$d$ convolutional layer with kernel size $1 \times L$ is applied on each input variable where $L$ is the length of the time interval of interest. The kernel window slides through the temporal domain with overlap. The convolutional weight is shared along the temporal domain and each input variable has its own dedicated feature extraction convolutional layer. In our model, we adopt a convolutional layer with $J$ filters so that a $J$-dimensional feature vector is extracted for each variable from each time interval. The convolutional layer encodes multivariate input sequence as follows:
\begin{equation}
    \label{eq:cnn_feature_extraction}
    \left\lbrace \mathbf{c}_{i,t} \right\rbrace = CNN_i\left( \mathbf{x}_i \right), \quad i=1, \dots, P
\end{equation}
where $\mathbf{c}_{i,t} \in \R^J$ is the feature vector for $\mathbf{x}_i$ extracted from the $t$-th time interval of interest, $t=1,\ldots,l$. Note that $l$, number of intervals, depends on the convolution kernel length $L$ and stride size, which is the number of time steps shifts over the input data matrix.

The convolution-based feature extraction yields features that incorporate the temporal pattern of values assumed by the input variables within a local context (determined by the convolution window). The attention mechanism applied to such features measures the importance of the targeted time interval, as opposed to specific time points. Thus, the  convolutional layers can learn to adapt to the dynamics of each input time series variable while ensuring that the attention scores are attached to the corresponding input variables. The multiple filters attend to different aspects of the signal and jointly extract a rich feature vector that encodes the relevant information from the time series in the time interval of interest. Note that for each variable, the convolution computation on each time interval can be carried out in parallel, as opposed to the sequential processing in canonical RNN models. Furthermore, the number of effective time points is significantly reduced by considering intervals as opposed to discrete time points. This also reduces the computational complexity for downstream attention mechanism. While we limit ourselves to the simple convolution structure described above, the LAXCAT architecture can accommodate more sophisticated e.g., dilated~\cite{oord2016wavenet} convolution structures for more flexible feature extraction from MTS data.

The feature extraction module accepts an input time series $\left\lbrace \mathbf{x}_1, \ldots, \mathbf{x}_T \right\rbrace$ and produces a sequence of feature matrices $\left\lbrace \mathbf{C}_1,\ldots,\mathbf{C}_{l} \right\rbrace$, where $\mathbf{C}_t \in \R^{P\times J}$. Each row in $\mathbf{C}_t$ stores the feature vector specific to each variable within time interval $t$ in the input sequence, i.e., $\mathbf{C}_t = [\mathbf{c}_{1,t}, \ldots, \mathbf{c}_{P,t}]^T$. The variable attention module (see below) considers the feature matrices at each time interval so as to obtain a local context embedding vector $\mathbf{h}_t, t=1,\ldots,l$, for each interval. The temporal attention module construct the summary embedding $\mathbf{z}$, which is used to encode the MTS data for classification. In the model, temporal attention measures the contribution of each time interval to the embedding whereas variable attention controls the extent to which each variable is important within each interval. The following subsections discuss the detail of the two attention modules.

\subsection{Variable Attention Module}
The variable attention module evaluates variable attention and constructs local context embedding. Specifically, the local context embedding is an aggregation of the feature vectors weighted by their relative importance measures within the specific time interval. The context vector $\mathbf{h}_t \in \R^J$ for the $t$-th time interval is obtained by
\begin{equation}
    \label{eq:context_vector_time}
    \mathbf{h}_t = \sum_{i=1}^P \alpha_{i_t}\mathbf{c}_{i,t}
\end{equation}
where $\alpha_{i_t}$ is the attention score for $\mathbf{c}_{i,t}$ and is equivalently the attention score dedicated to variable $\mathbf{x}_i$ in the $t$-th time interval.

To precisely evaluate the importance of each variable, we use a feed forward network to learn the attention score vectors $\mathbf{a}_t = \left[ \alpha_{1_t}, \ldots, \alpha_{P_t} \right], t=1,\ldots,l$. The network can be characterized by
\begin{equation}
    \label{eq:feature_attention}
    \begin{aligned}
        &\mathbf{a}_t = softmax \left( \mathbf{s}_t \right)\\
        &\mathbf{s}_t = \sigma_2 \left( \sigma_1 \left( W_1^{\left( V \right)} \mathbf{C}_t + B_1^{\left( V \right)} \right)W_2^{\left( V \right)} + B_2^{\left( V \right)} \right)
    \end{aligned}
\end{equation}
where $W_1^{\left( V \right)} \in \R^{1 \times P}$, $W_2^{\left( V \right)} \in \R^{J \times P}$, $B_1^{\left( V \right)} \in \R^{1 \times J}$, $B_2^{\left( V \right)} \in \R^{1 \times P}$ are the model parameters, and $\sigma_1 \left( \cdot \right)$, $\sigma_2 \left( \cdot \right)$ are non-linear activation functions such as  tanh, ReLU among others. In the feature extraction stage in Sec.~\ref{sec:feature_extraction}, each time series input variable is processed independently and the correlations among the variables have not been considered. The input data to the variable attention network is the feature matrix $\mathbf{C}_t$ which contains all feature vectors in time interval $t$. The attention network considers the multivariate correlation and distributes attention weights to each variable so as to maximize the predictive performance. Note that the local context embeddings in different intervals can be constructed independently of each other (and hence processed in parallel). In addition, the parameters of the variable attention module are shared among all intervals to ensure parsimony with respect to the model parameters.

The preceding process yields local context embeddings for each of the time intervals by considering the relative variable importance. The result is a context matrix $\mathbf{H} = \left[ \mathbf{h}_1, \ldots, \mathbf{h}_{l} \right]^T \in \R^{l \times J}$ consisting of the context vector at each interval. In the next subsection, we describe how to compose the summary embedding of the MTS instance using the temporal attention mechanism.

\subsection{Temporal Attention Module}
The goal of temporal attention module is to identify key segments of signals which contain information that can discriminate between classes. The summary vector $\mathbf{z}$ is composed by aggregating the context embedding vectors weighted by their relative temporal contribution as follows:
\begin{equation}
    \label{eq:summary_vector}
    \mathbf{z} = \sum_{t=1}^{l} \beta_t \mathbf{h}_t
\end{equation}
where $\beta_t$ is the temporal attention score for the context vector $\mathbf{h}_t$ and it quantifies the contribution of the information carried in interval $t$. Similarly, the temporal attention module is instantiated by a feed-forward network. The vector of temporal attention scores $\mathbf{b} = [\beta_1,\ldots,\beta_l]$ is learned using the following procedure:
\begin{equation}
    \label{eq:temporal_attention}
    \begin{aligned}
        &\mathbf{b} = softmax \left( \mathbf{u} \right)\\
        &\mathbf{u} = \sigma_2 \left( \sigma_1 \left( W_1^{\left( T \right)} \mathbf{H} + B_1^{\left( T \right)} \right)W_2^{\left( T \right)} + B_2^{\left( T \right)} \right)
    \end{aligned}
\end{equation}
where $W_1^{\left( T \right)} \in \R^{1 \times l}$, $W_2^{\left( T \right)} \in \R^{J \times l}$, $B_1^{\left( T \right)} \in \R^{1 \times J}$, and $B_2^{\left( T \right)} \in \R^{1 \times l}$ are model parameters, and $\sigma_1 \left( \cdot \right)$, $\sigma_2 \left( \cdot \right)$ are non-linear activation functions. The input to the temporal attention module is the entire context matrix $\mathbf{H}$. The module takes into account the correlations among time intervals and the predictive performance of each interval to distribute attention scores.

This concludes the description of the three modules of LAXCAT architecture. To summarize, the convolutional feature extraction module extracts a rich set of features from time series data.  The variable and temporal attention modules, construct an embedding of the MTS data to be classified by attending to the relevant variables and time intervals.

\subsection{Learning LAXCAT}
Given the encoding $\mathbf{z}$ which captures the important variables and time intervals of the input multivariate time series sequence, we can predict the class label of the sequence as follows:
\begin{equation}
    \label{eq:prediction}
    y = f(\mathbf{z}; \mathbf{W})
\end{equation}
where $\mathbf{W}$ is the weights of $f(\cdot)$, a fully connected feed-forward network\footnote{Although here we focus on classifying MTS data, the LAXCAT framework can be readily applied to forecasting and other related tasks by choosing an appropriate $f(\cdot)$ .}.

\begin{figure}
    \centering
    \includegraphics[width = 0.35\textwidth]{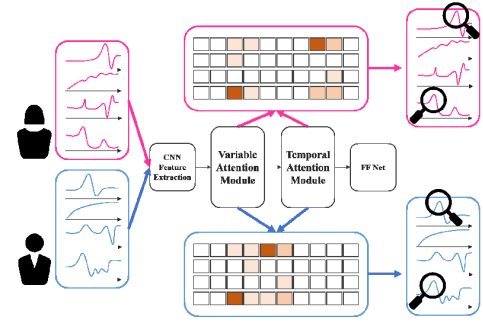}
    \caption{Explainability of the LAXCAT model.}
    \label{fig:interpret_att}
\end{figure}

Given a set of training instances $\left\lbrace (\mathbf{X}_1,y_1),\ldots,(\mathbf{X}_N,y_N) \right\rbrace$, the parameters $\Theta$ of the variable and temporal attention modules and the classifier network can be jointly learned by optimizing the following objective function:
\begin{equation}
    \label{eq:model_obj}
    \min_{\Theta} \frac{1}{N} \sum_{i=1}^N \mathcal{L}(\mathbf{X}_i, y_i; \Theta) + \alpha \|\Theta\|_F^2
\end{equation}
where $\|\Theta\|_F^2$ is the Frobenius norm on the weights to alleviate overfitting and $\alpha$ is a scalar that control the effect of the regularization term. In this study,  $\mathcal{L}$ is chosen to be the cross entropy loss function. The resulting objective function is smooth and differentiable allowing the objective function to be minimized using standard gradient back propagation update of the model parameters. We used the Adam optimizer~\cite{kingma2014adam} to train the model parameters where the hyperparameters of the Adam optimizer are set to their default values.
\vskip 0.5em
\noindent{}\textbf{Explainability of LAXCAT}. Recall that LAXCAT is designed to accurately classify MTS data and also facilitate instance-level explanation of the predicted class label for each MTS instance by identifying the key time intervals and variables that contribute to the classification. Given an input sequence, we can extract two attention measures from the model, namely temporal attention scores and variable attention scores. A summary of the importance of each variable in each interval is given by the product of the two attention scores,
\begin{equation}
    \label{eq:intrepretation}
    JointAtt_{i,t} = \alpha_{i_t}\times \beta_t
\end{equation}
for $i=1,\ldots,P$, $t = 1,\ldots,l$ (See Figure~\ref{fig:interpret_att}). These results can then be compared against domain knowledge or used to guide further experiments. 

\section{Experiments and Results}
We proceed to describe our experiments aimed at  evaluating the performance LAXCAT in terms of the accuracy of MTS classification as well as the explainability of the resulting classifications.

\begin{table}[t]
\small
    \caption{A summary of the datasets.}
    \label{tbl:datasummary}
    \vskip -1em
    \begin{tabular}{l|cccc}
        Dataset    & \# Var. ($P$) & \# Time Points & \# Classes & \# Samples \\ \hline
        PM2.5 w/   & 8            & 24             & 6          & 1013         \\
        PM2.5 w/o  & 7            & 24             & 6          & 1013         \\
        Seizure    & 23           & 1025           & 2          & 272          \\
        Motor Task & 15           & 641            & 2          & 405         
    \end{tabular}
\end{table}

\subsection{Datasets}
\begin{figure}[t]
    \centering
    \subfigure[Seizure data]{ 
        \includegraphics[width=0.225\textwidth]{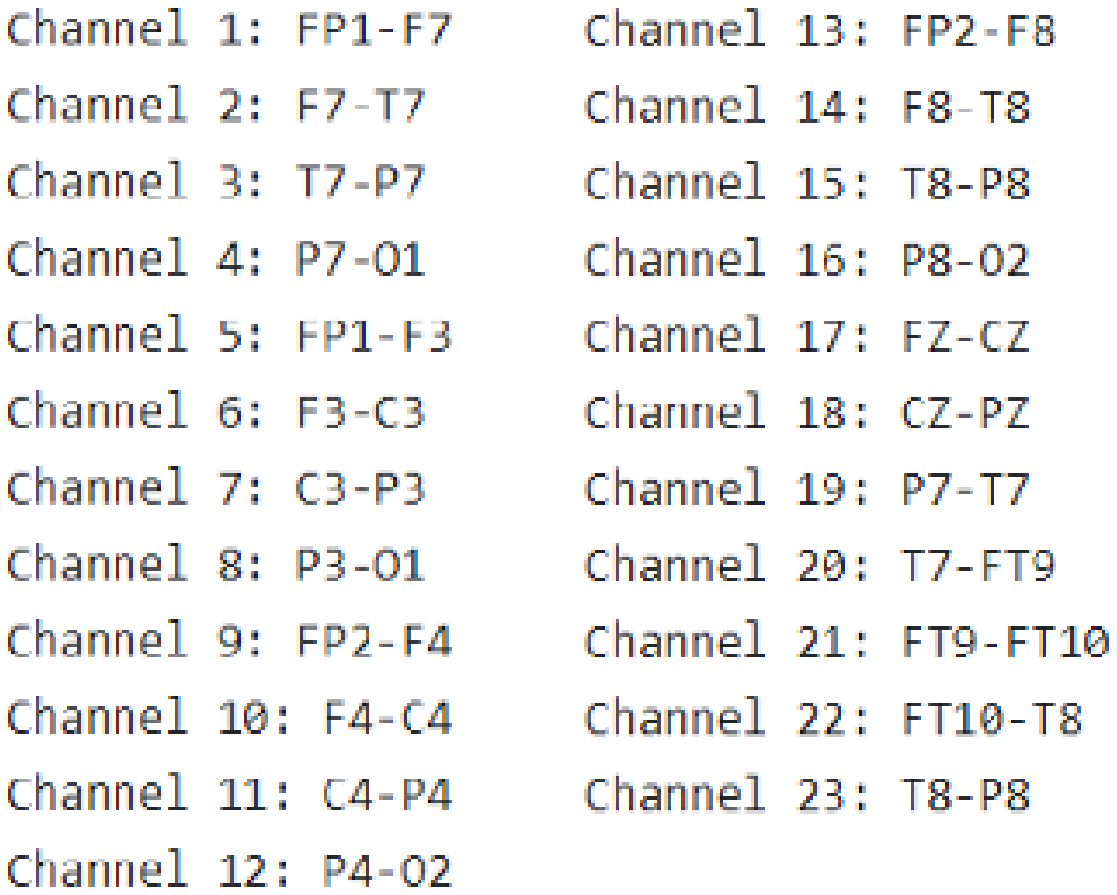}%
        \label{fig:ch_location:seizure}%
    }
    \subfigure[Movement data]{
        \includegraphics[width=0.225\textwidth]{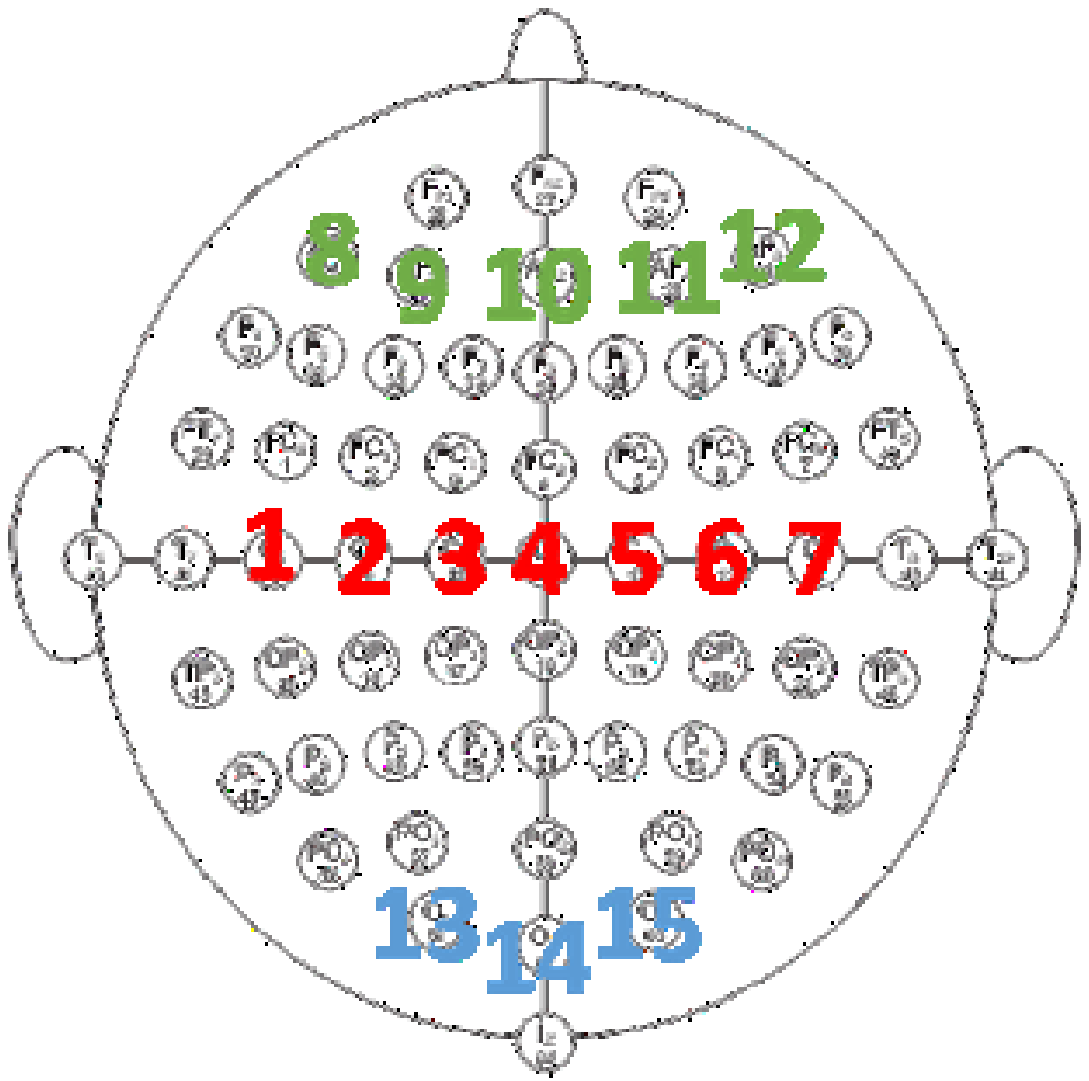}%
        \label{fig:ch_location:mi}%
    }
    \vskip -1em
    \caption{Variable number and EEG location correspondence for Seizure and Movement data sets.}
    \vspace{-10pt}
    \label{fig:ch_location}
\end{figure}

We used three publicly available real-world multivariate time series data sets and a summary of the data sets is provided in Table~\ref{tbl:datasummary}.
\begin{itemize}[leftmargin=*]
\item\textbf{PM2.5 data set}~\cite{liang2015assessing} contains hourly PM2.5 value and the associated meteorological measurements in Beijing City of China. Given the measurements on one day, the task is to predict the PM2.5 level on the next day at 8 am, during the peak commute. The PM2.5 value is categorized into six levels according to the United States Environmental Protection Agency standard, i.e., good, moderate, unhealthy for sensitive, unhealthy, very unhealthy, and hazardous. We arranged this data set into two versions, the first one contains PM2.5 recordings as one of the covariates, called \textbf{PM2.5 w/}, and the second one excludes PM2.5 recordings, denoted as \textbf{PM2.5 w/o}. Aside from PM2.5 values, the meteorological variables include dew point, temperature, pressure, wind direction, wind speed, hours of snow and hours of rain. We keep the measurements for  weekdays and exclude the measurements for weekends yielding a data set of 1013 MTS instances in total.

\item\textbf{Seizure data set}~\cite{shoeb2009application,goldberger2000physiobank}  consists of electroencephalogram (EEG) recordings from pediatric subjects with intractable seizures collected at the Children's Hospital Boston. EEG signals at 23 positions, as shown in Figure~\ref{fig:ch_location:seizure}, according to the international 10-20 system, were recorded at 256 samples per second with 16-bit resolution. Each instance is a four second recording containing either seizure attack period or non-seizure period.

\item\textbf{Movement data set}~\cite{schalk2004bci2000,goldberger2000physiobank} consists of EEG recordings of subjects opening and closing left or right fist. EEG signals were recorded at 160 samples per second and 15 electrode locations were used in this study, covering the central-parietal, frontal and occipital regions as shown in Figure~\ref{fig:ch_location:mi}. Each instance contains 4 second recordings and the subjects were at rest state during the first two seconds and performed the fist movement during the latter two seconds. The task is to distinguish between left and right fist movement based on the 15-channel EEG recordings.
\end{itemize}

\subsection{Baseline Methods and Evaluation Setup}
We compare the classification performance of LAXCAT with representative and state-of-the-art baselines:

\begin{itemize}[leftmargin=*]
\item \textbf{$k$NN-DTW}~\cite{friedman2001elements,muller2007dynamic} is the dynamic time warping (DTW) distance measure combined with $k$-nearest neighbor ($k$NN) classifier. DTW provides a similarity score between two time series by warping the time axes of the sequences to achieve alignment. The classification phase is carried out by $k$NN classifier.
\item\textbf{LR} is the logistic regression classifier. For multivariate time series input, we concatenate all variables and the input to the LR model is a multivariate vector.
\item\textbf{LSTM}~\cite{hochreiter1997long} is the long short-term memory recurrent neural network. An LSTM network with one hidden layer is adopted to learn an encoding from multivariate time series data and the classification phase is carried out by a feed forward neural network.
\item\textbf{DARNN}~\cite{qin2017dual} is a dual attention RNN model. It uses an encoder-decoder structure where the encoder is applied to learn attentions and the decoder is adopted for prediction task.
\item\textbf{IMV-LSTM}~\cite{guo2019exploring} is the interpretable multivariate LSTM model. It explores the structure of LSTM networks to learn variable-wise hidden states. With hidden states, a mixture attention mechanism is exploited to model the generative process of the target.
\end{itemize}






We implemented the proposed model and deep learning baseline methods with PyTorch~\cite{paszke2019pytorch}. We used the Adam optimizer~\cite{kingma2014adam} to train the networks with default parameter settings and the mini-batch size is 40. The number of filters in the feature extraction step is chosen from $\left\lbrace 8,16,32 \right\rbrace$. The kernel size $L$ is selected from $\left\lbrace 2,3,5 \right\rbrace$, $\left\lbrace 16,32,64 \right\rbrace$, $\left\lbrace 16,32,64 \right\rbrace$ in PM2.5, Seizure, Motor Task respectively, and the stride size is set to $50\%$ of kernel size. For the number of hidden nodes in the classifier feed forward network, we conduct grid search over $\left\lbrace 8,16,32 \right\rbrace$. The coefficient for the regularization term is chosen from $\left\lbrace 0.001,0.01,0.1 \right\rbrace$. In the case of the $k$NN-DTW method, $k$ is set to 1, yielding an one nearest neighbor classifier. In the case of the LSTM baseline, the number of hidden nodes is selected from $\left\lbrace 8,16,32,64 \right\rbrace$. For IMV-LSTM, we implemented IMV-Tensor as it was reported to perform better~\cite{guo2019exploring}. The parameter selection for the baseline methods DARNN and IMV-LSTM follows the guidelines provided in the respective papers. In the training phase, 70\% of the samples are used to train the models and 15\% of the samples are for validation. The remaining 15\% are used as test set. We repeat the experiment five times and report the average performance.

\subsection{Performance of LAXCAT}

\noindent{\bf Classification Accuracy}.
We compared LAXCAT with the baseline methods on MTS classification using the different benchmark data sets described above and report the results in Table~\ref{tbl:classify_accuracy}. The results of our experiments show that deep learning-based methods outperform the other two simple baseline methods, 1NN-DTW and LR. LR mostly outperforms 1NN-DTW with the only exception on Seizure data set. Among the deep learning based methods, those equipped with an attention mechanism achieve better classification accuracy than the canonical LSTM model. Among the attention based deep neural network  models, LAXCAT outperforms the other two attention based deep neural network baselines. We further note that, in the case of the two PM2.5 data sets, perhaps not surprisingly,  all models consistently make better future PM2.5 value predictions when past PM2.5 value recordings are included as an input .

\noindent
\textbf{Time Complexity.} We also compare the computational complexity of deep learning based methods in terms of run-time per training iteration and run-time per testing iteration. As reported in Table~\ref{tbl:runtime}, the LSTM baseline does not include any attention mechanism to track variable and temporal importance and hence takes the least amount of run-time in each training and test iteration  across all the data sets. Among the attention based deep neural network models, LAXCAT has the shortest run-time. The difference in execution time between LAXCAT and the two baselines is quite substantial in the case of Seizure and Motor Task data sets, due to the  lengths of the time series in question: Each sequence in Seizure data set contains 1025 sampling points while sequences in Motor Task contain 641 sampling points. DARNN and IMV-LSTM evaluate time point-based attention, which places a greater computational burden compared to the time interval-based attention in LAXCAT . We further note in contrast to LSTM based methods which are inherently sequential, many aspects of LAXCAT are parallelizable.

\begin{table*}[t]
    \caption{Classification results (Accuracy$\pm$std) of different algorithms on the four data sets}
    \label{tbl:classify_accuracy}
    \vskip -1em
    \begin{tabular}{l|cccccc} \hline
        Dataset    & 1NN-DTW      & LR           & LSTM         & DARNN        & IMV-LSTM     & Proposed     \\ \midrule
        PM2.5 w/   & 36.05 $\pm$ 3.24 & 38.29 $\pm$ 0.86 & 40.40 $\pm$ 1.89 & 41.19 $\pm$ 2.89 & 48.16 $\pm$ 3.30 & $\mathbf{50.66 \pm 4.58}$ \\
        PM2.5 w/o  & 30.39 $\pm$ 3.85 & 35.92 $\pm$ 2.11 & 37.06 $\pm$ 3.04 & 38.98 $\pm$ 5.43 & 39.34 $\pm$ 4.40 & $\mathbf{45.53 \pm 6.20}$ \\
        Seizure    & 53.66 $\pm$ 7.72 & 52.20 $\pm$ 5.88 & 70.24 $\pm$ 2.67 & 71.31 $\pm$ 2.78 & 72.19 $\pm$ 2.78 & $\mathbf{76.59 \pm 4.08}$ \\
        Movement & 53.44 $\pm$ 5.13 & 71.80 $\pm$ 8.31 & 75.32 $\pm$ 3.26 & 83.28 $\pm$ 1.37 & 84.09 $\pm$ 1.12 & $\mathbf{87.21 \pm 1.37}$  \\ \bottomrule      
    \end{tabular}
\end{table*}

\begin{table}[]
    \caption{Run-time (in seconds) comparison. Run-time for each training iteration is reported at the top half. Run-time for the testing phase is reported at the bottom half.}
    \label{tbl:runtime}
    \vskip -1.5em
    \begin{tabular}{r|ccc} \hline
        Dataset  & PM2.5w/ & Seizure & Movement \\ \toprule
        LSTM (Train)     & 0.5     & 4       & 2.8        \\
        DARNN (Train)    & 4.8     & 430     & 218        \\
        IMV-LSTM (Train) & 3.3     & 430     & 150        \\
        Ours (Train)     & 1.4     & 4.5     & 3.5        \\ \midrule
        LSTM (Test)     & 0.001   & 0.01    & 0.01       \\
        DARNN (Test)    & 0.08    & 24      & 13         \\
        IMV-LSTM (Test) & 0.06    & 20      & 4.3        \\
        Ours (Test)     & 0.02    & 0.03    & 0.03       \\ \bottomrule
    \end{tabular}
\end{table}


\subsection{Explaining the LAXCAT Predictions}
We proceed to describe several case studies designed to evaluate the effectiveness of LAXCAT in producing useful explanations of its classification of each MTS instance. For qualitative analysis, we report the meaningful variables and time intervals identified by the attention mechanism and compare them with domain findings in related literature. To quantitatively assess the effectiveness of the allocation of attention, we define the attention allocation measure (AAM)
\begin{equation}
    \label{eq:att_allocation_measure}
    AAM = \frac{\textnormal{Amount of attention allocated correctly}}{\textnormal{Total amount of attention}} \times 100\%
\end{equation}
This measure is only applied to the cases where correct allocation of attention is properly defined, i.e. a solid understanding of important variable and time interval is present.

\subsubsection{Case Study I: Synthetic Data}
To thoroughly examine the attention mechanism, we constructed a synthetic data set that reflects  concrete prior knowledge regarding the key variables as well as the time intervals that determine class labels. This synthetic data set consists of 3 time series variables, i.e. $\mathbf{x}^{(t)}_1 = \cos{(2\pi t)} + \epsilon$, $\mathbf{x}^{(t)}_2 = \sin{(2\pi t)} + \epsilon $, and $\mathbf{x}^{(t)}_3 = \exp{(t)} + \epsilon$, where $\epsilon$ is Gaussian noise and $t$ takes value from a vector of 50 linearly equally spaced points between 0 and 1. To generate two classes, we randomly select half of the instances and manipulate the first variable $\mathbf{x}^{(t)}_1$ by adding a square wave signal to the raw sequence. The square wave is controlled by three random variables, the starting point, the length, and the magnitude of the square wave. We treat instances with square wave as positive and instances without square wave as negative. For the synthetic data, we define correct attention allocation as the attention assigned to variable 1 within the interval of the square wave. The AAM scores on the synthetic data are reported in Table~\ref{tbl:aam_score}. From the table, we observe that LAXCAT outperforms DARNN and IMV-LSTM the baseline attention based deep neural network models, suggesting that that LAXCAT can better identify important variables as well as the relevant time intervals that define the pattern of values that influence the class labels.  

\begin{figure}[t]
    \centering
    \subfigure[Positive Sample]{ 
        \includegraphics[width=0.225\textwidth]{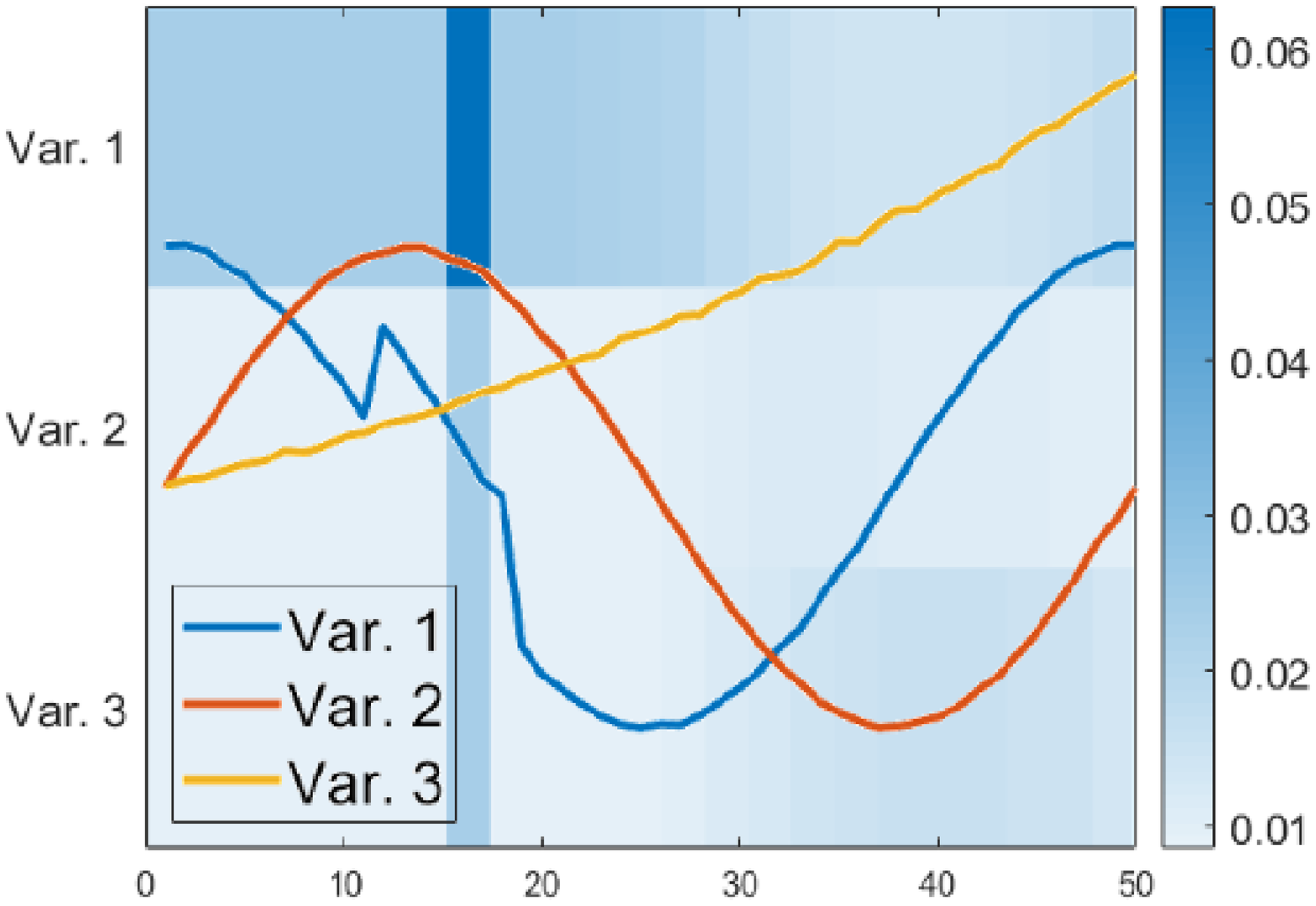}%
        \label{att_syn:subj1}%
    }
    \subfigure[Negative Sample]{
        \includegraphics[width=0.225\textwidth]{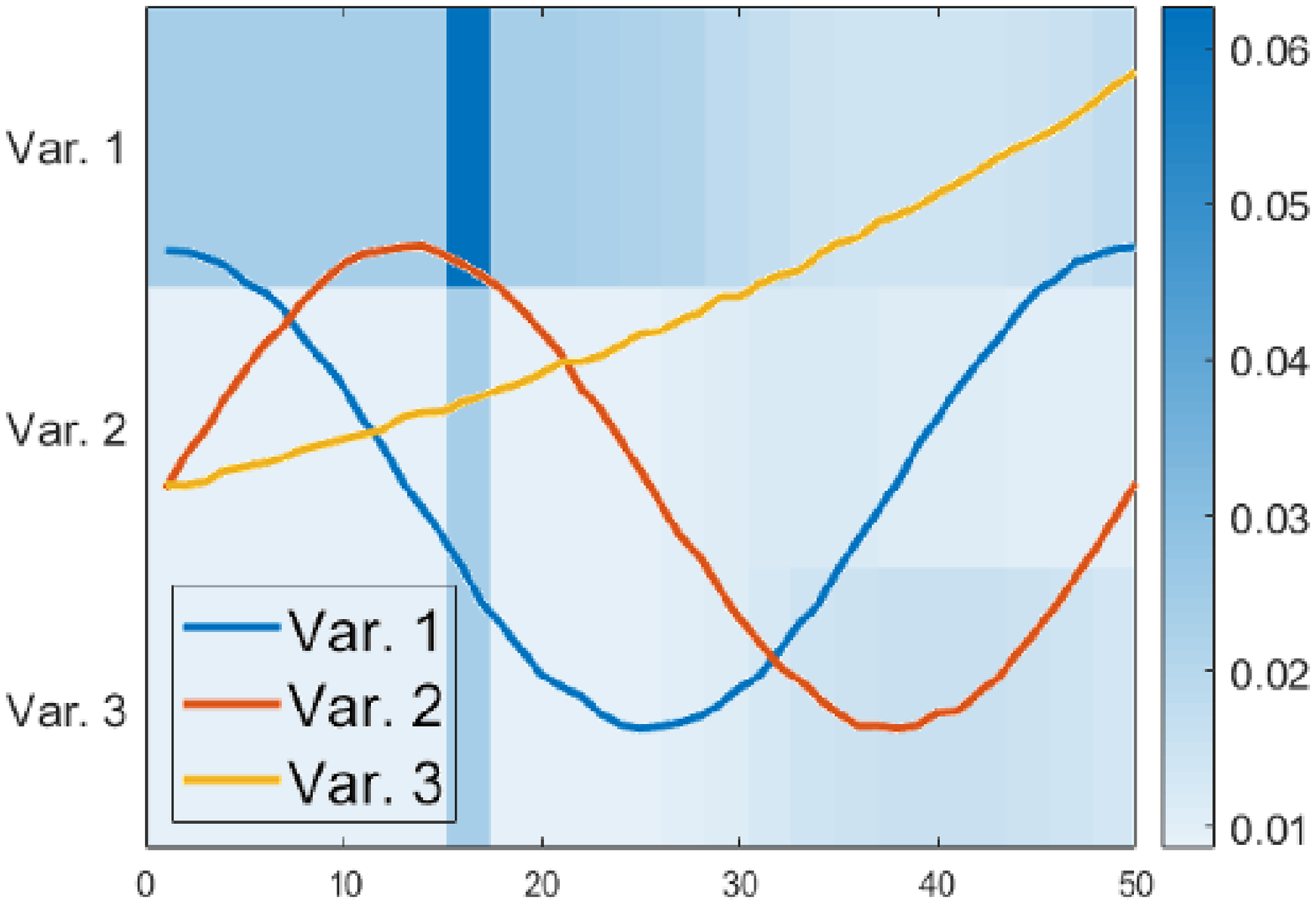}%
        \label{att_syn:subj2}%
    }\\
    \vskip -1.5em
    \caption{Positive and negative synthetic examples are drawn in solid lines and the heat maps of the attention allocation are depicted in the background. The attention for variable 1 is located in the top row, variable 2 in the center row, and variable 3 in the bottom row. (Best viewed in color)}
    \label{fig:att_syn}
    \vskip -1em
\end{figure}

We give an illustration of positive instance and negative instance with the attention allocation by our proposed method in Figure~\ref{fig:att_syn}. We observe that the proposed model distributes most of its attention to variable 1, specifically, in the interval that covers  the location of the square wave in both positive and negative class instances.

\begin{table}[]
    \caption{AAM score on Synthetic data and Motor Task.}
    \label{tbl:aam_score}
    \vskip -1em
    \begin{tabular}{r|ccc} \hline
        Dataset    & DARNN & IMV-LSTM & Ours  \\ \toprule
        Synthetic  & 5.42\%    &    7.91\%      &  10.93\%     \\
        Motor Task & 19.91\% & 22.08\%    & 24.17\%         \\ \bottomrule
    \end{tabular}
    \vspace{-10pt}
\end{table}

\subsubsection{Case Study II: PM2.5}
\begin{figure}[t]
    \centering
    \vskip -1em
    \subfigure[PM2.5 w/]{ 
        \includegraphics[width=0.225\textwidth]{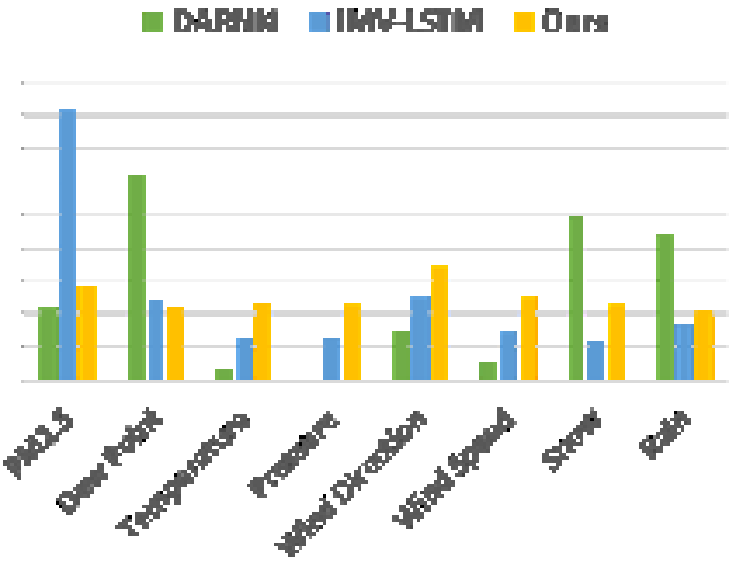}
        \label{fig:att_pm25:with}
    }
    \subfigure[PM2.5 w/o]{
        \includegraphics[width=0.225\textwidth]{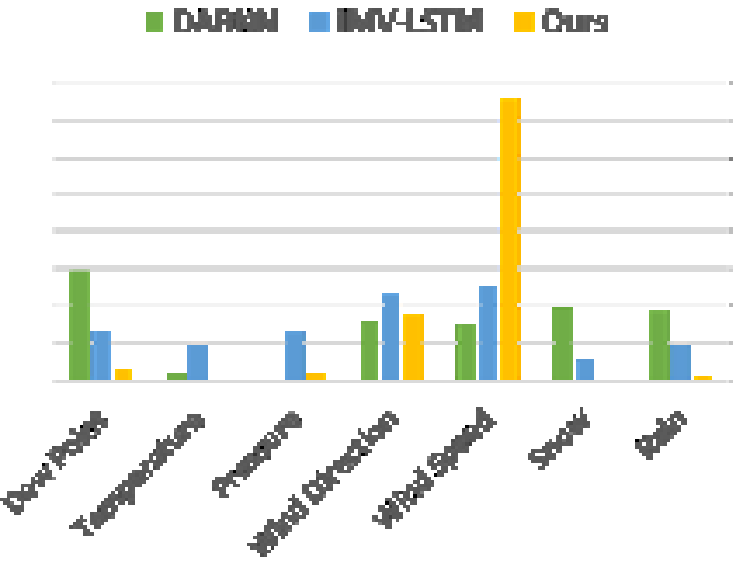}
        \label{fig:att_pm25:without}
    }\\
    \vskip -1.5em
    \caption{Average attention allocation on the PM25 datasets.}
    \vskip -1.5em
    \label{fig:att_pm25}
\end{figure}

PM2.5 value is the concentration of particles with a diameter of 2.5 micrometers or less suspended in air. Studies have found a close link between exposure to fine particles and premature death from heart and lung disease~\cite{zanobetti2000airborne,franklin2008role}. We report the attention learning results in Figure~\ref{fig:att_pm25}. Variable-wise speaking, as shown in Figure~\ref{fig:att_pm25}, when past PM2.5 recordings are available for future prediction, IMV-LSTM ranks PM2.5 as the most important variable, and the LAXCAT method ranks it as the second most important variable. DARNN consistently selects dew point, snow, and rain as important predictive variables. When PM2.5 value is not available, wind speed, wind direction and pressure are high ranked by IMV-LSTM, which is consistent with that reported in~\cite{guo2019exploring}. LAXCAT attends to wind direction and wind speed besides PM2.5 value. According to~\cite{pu2011effect}, wind direction and speed are critical factors that affect the amount of pollutant transport and dispersion between Beijing and surrounding areas.

\begin{figure}[t]
    \centering
    \subfigure[Left hand 1]{
        \includegraphics[width=0.11\textwidth]{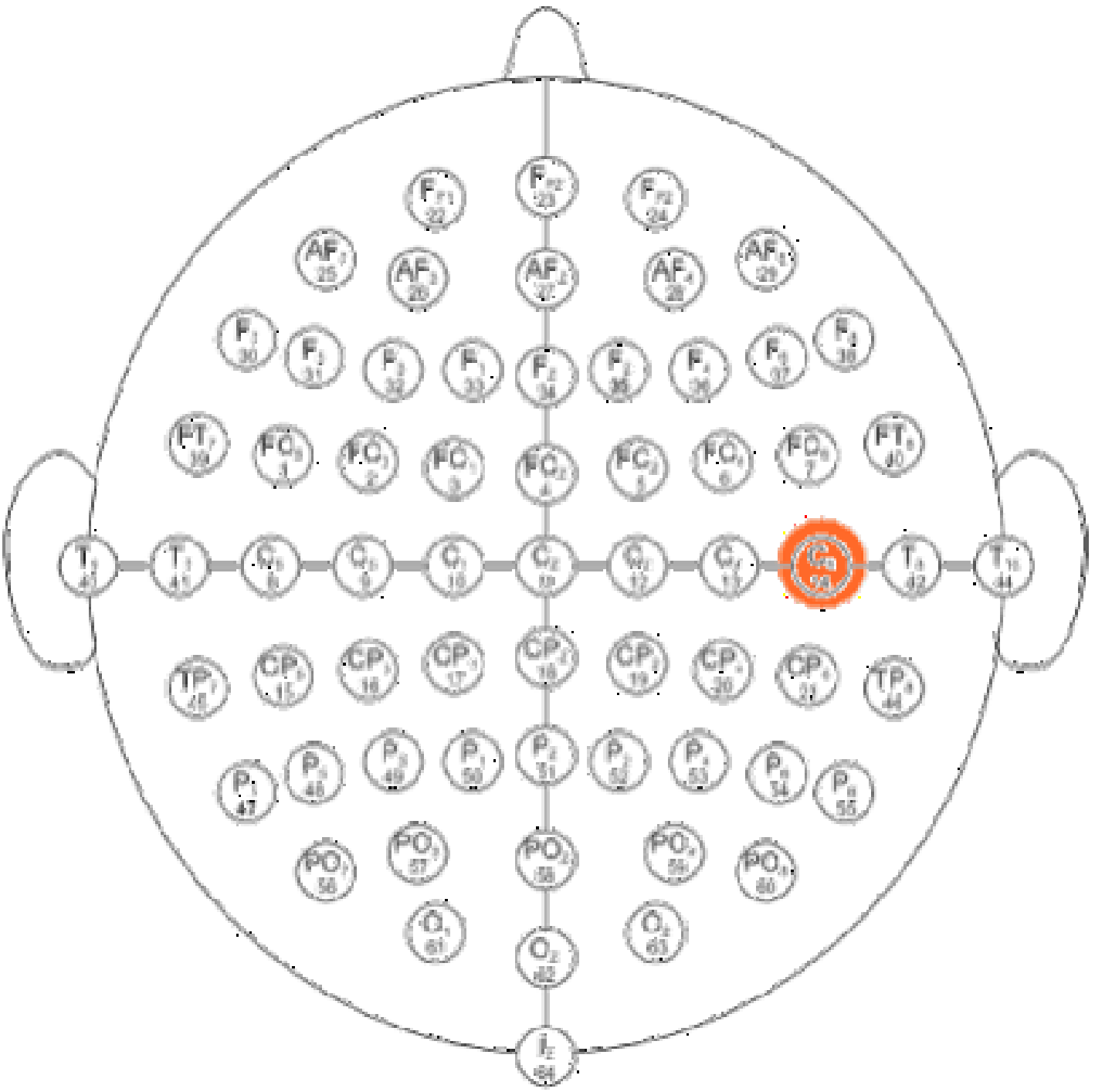}%
        \label{fig:att_mi:left1}%
    }
    \subfigure[Right hand 1]{ 
        \includegraphics[width=0.11\textwidth]{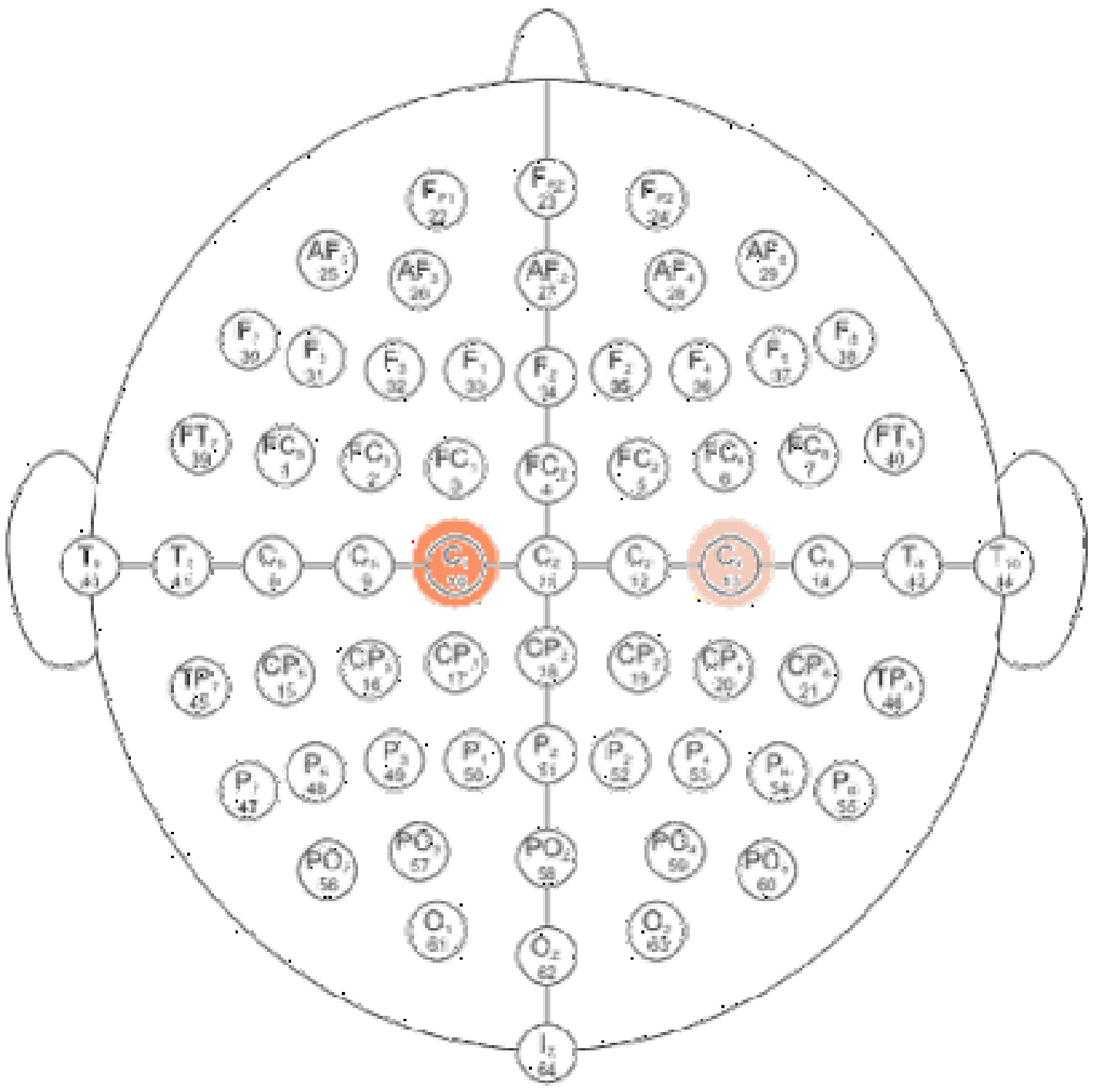}%
        \label{fig:att_mi:right1}%
    }
    \subfigure[Left hand 2]{
        \includegraphics[width=0.11\textwidth]{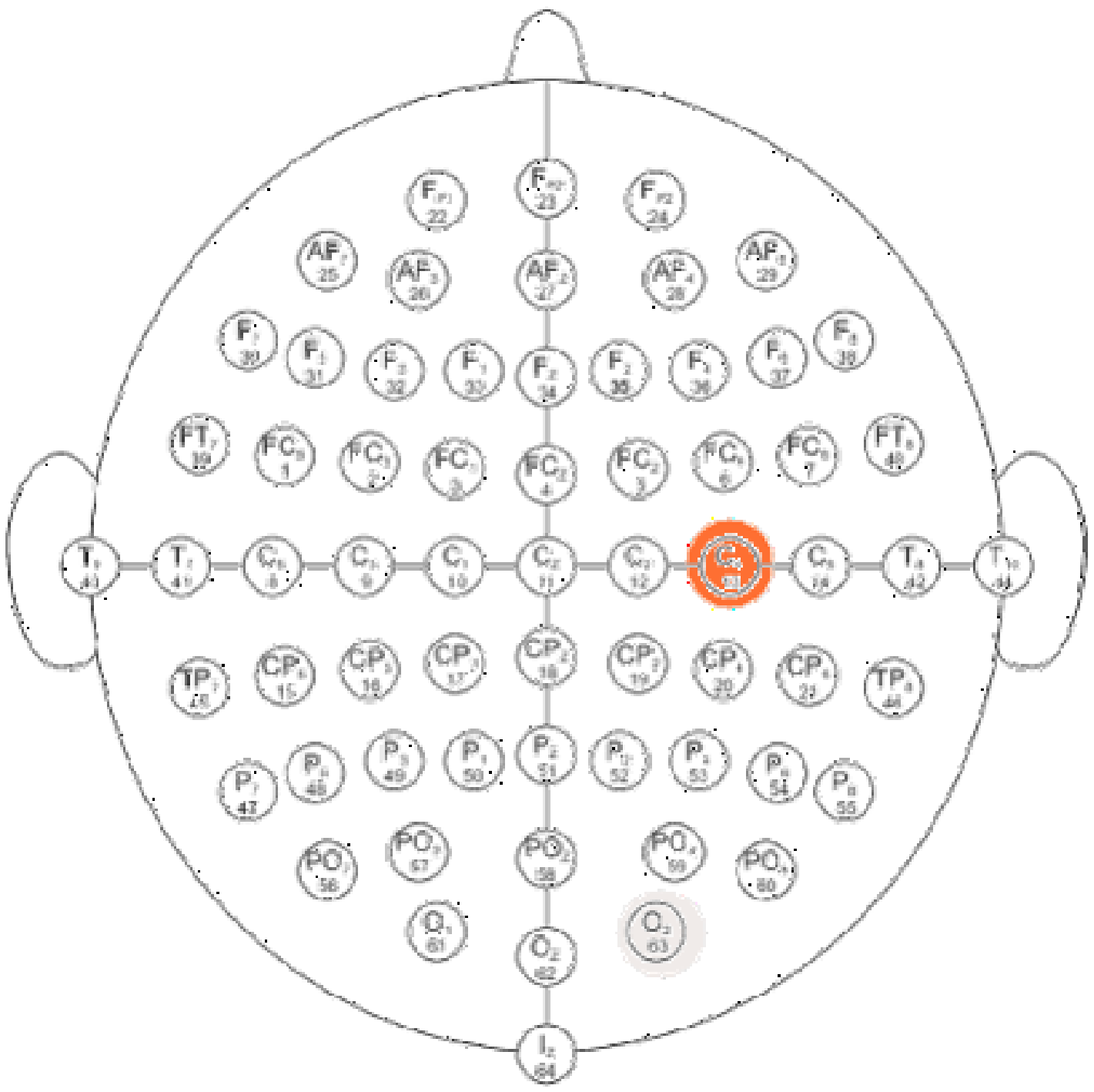}%
        \label{fig:att_mi:left2}
    }
    \subfigure[Right hand 2]{ 
        \includegraphics[width=0.11\textwidth]{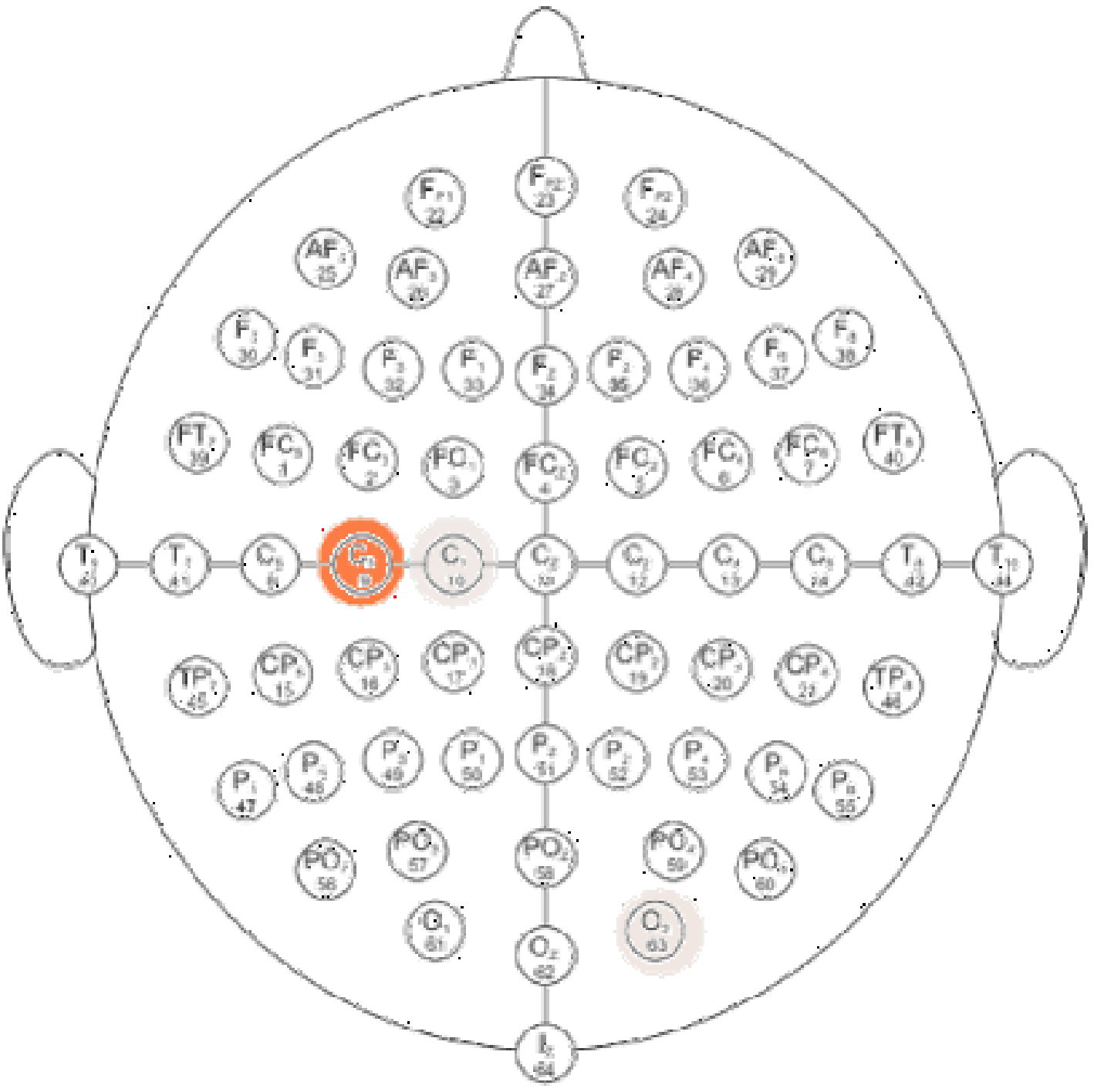}%
        \label{fig:att_mi:right2}%
    }
    \subfigure[Temporal attention]{
        \includegraphics[width=0.42\textwidth]{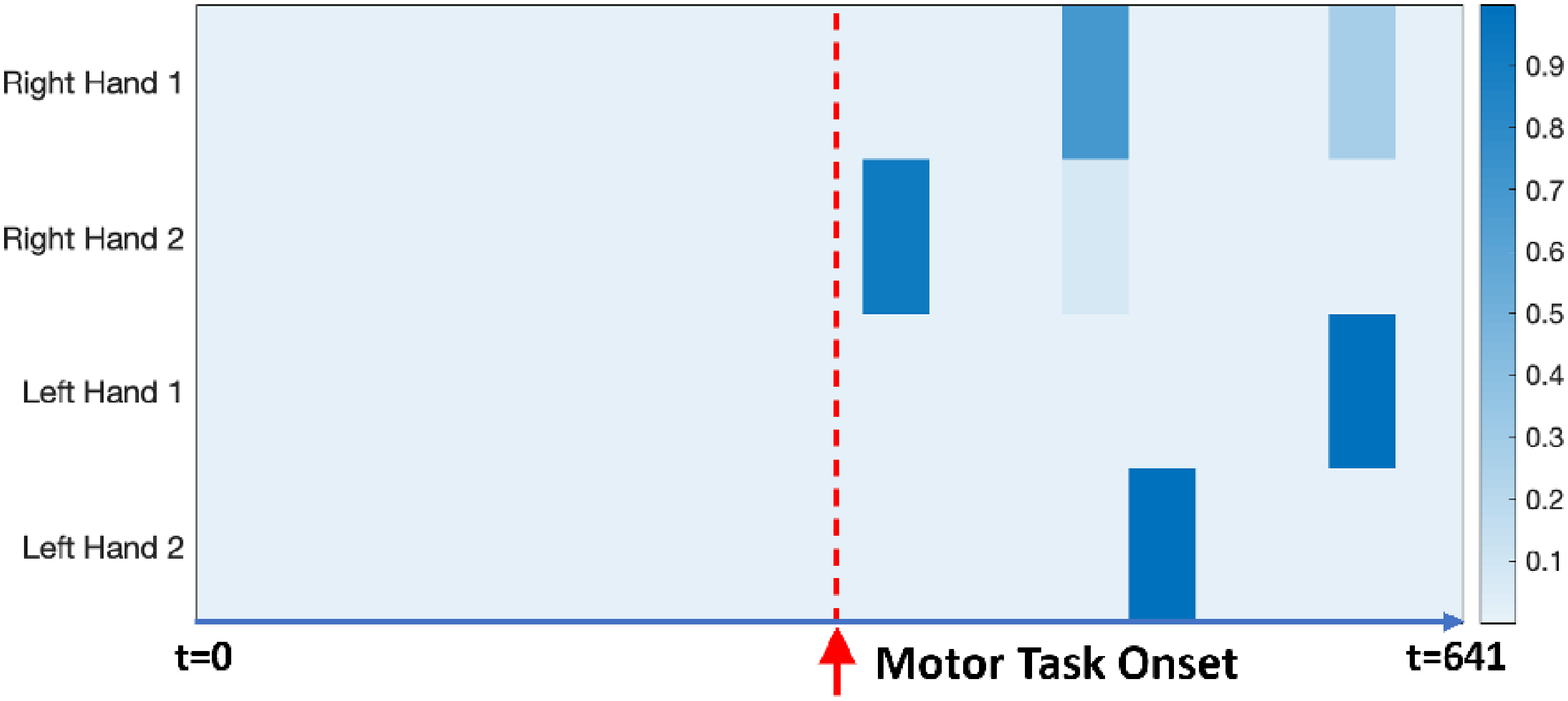}%
        \label{fig:att_mi:temporal}
    }
    \vskip -1.5em
    \caption{Important channels identified by LAXCAT on the Movement data set. The darker the color, the more attention allocated at the location. In the figure, all heads are facing top. Temporal attention result is reported at the bottom.}
    \vspace{-15pt}
    \label{fig:att_mi}
\end{figure}

\subsubsection{Case Study III: Movement Data}
On the Movement data set, we use EEG recordings to distinguish whether the subject is moving the left or right hand. A subset of attention results are reported in Figure~\ref{fig:att_mi}. Extensive research has shown that the motor cortex is involved in the planning, control, and execution of voluntary movements~\cite{sanes1995shared,petersen2012motor,conway1995synchronization}. The motor cortex, located in the rear portion of the frontal lobe, is closest to the locations of variables 1 to 7 in our empirical analysis. In Figures~\ref{fig:att_mi:left1}-~\ref{fig:att_mi:right2}, the heatmaps of accumulated attention in the entire time period from 2 subjects are depicted. We observe that most attention is distributed to the channels around the central region, namely the $C_1$, $C_3$, $C_4$, and $C_6$ channels. On the two left hand movement examples, i.e. Figure~\ref{fig:att_mi:left1},~\ref{fig:att_mi:left2}, the proposed model allocates attention to the right brain. On the contrary, the proposed model assigns most attention to the left brain during right hand movements as shown in Figure~\ref{fig:att_mi:right1},~\ref{fig:att_mi:right2}. This observation is in line with the current theory of contralateral brain function, which states that the brain controls the opposite side of the body. For temporal attention, as reported in Figure~\ref{fig:att_mi:temporal}, the proposed model lays most emphasis on the time intervals later to the motor task onset. For qualitative analysis, we define correct attention assignment as the attention distributed to channels $C_5,C_3,C_1,C_z,C_2,C_4,C_6$ in the time period after motor task onset. As reported in Table~\ref{tbl:aam_score}, LAXCAT assigns about 24\% of attention to the target zone as compared to around 20\% for DARNN and 22\% for IMV-LSTM. 

\begin{figure}[t]
    \centering
    \subfigure[8 hidden nodes]{
        \includegraphics[width=0.15\textwidth]{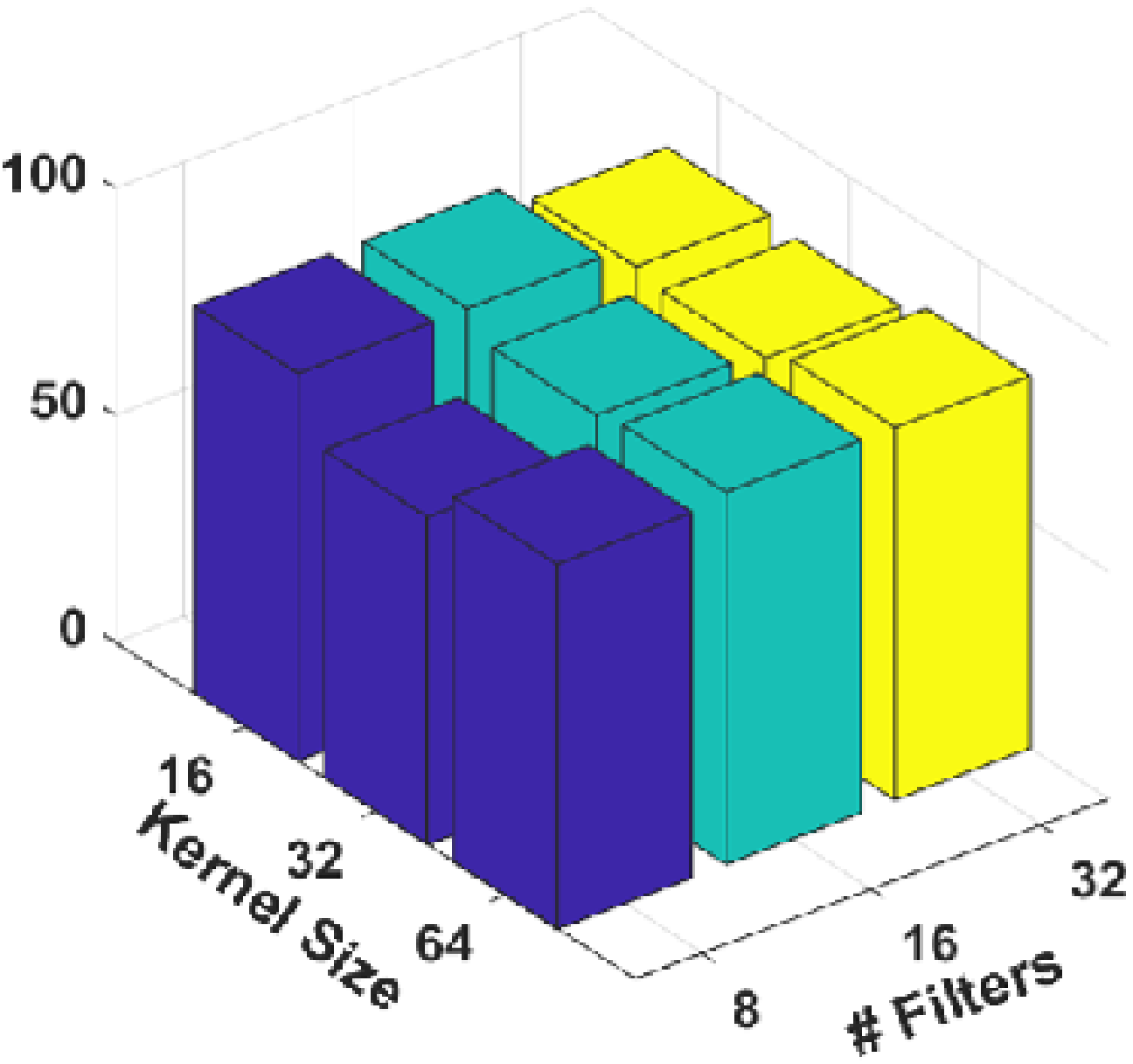}%
        \label{fig:par_mi_cat:hd8}%
    }
    \subfigure[16 hidden nodes]{ 
        \includegraphics[width=0.15\textwidth]{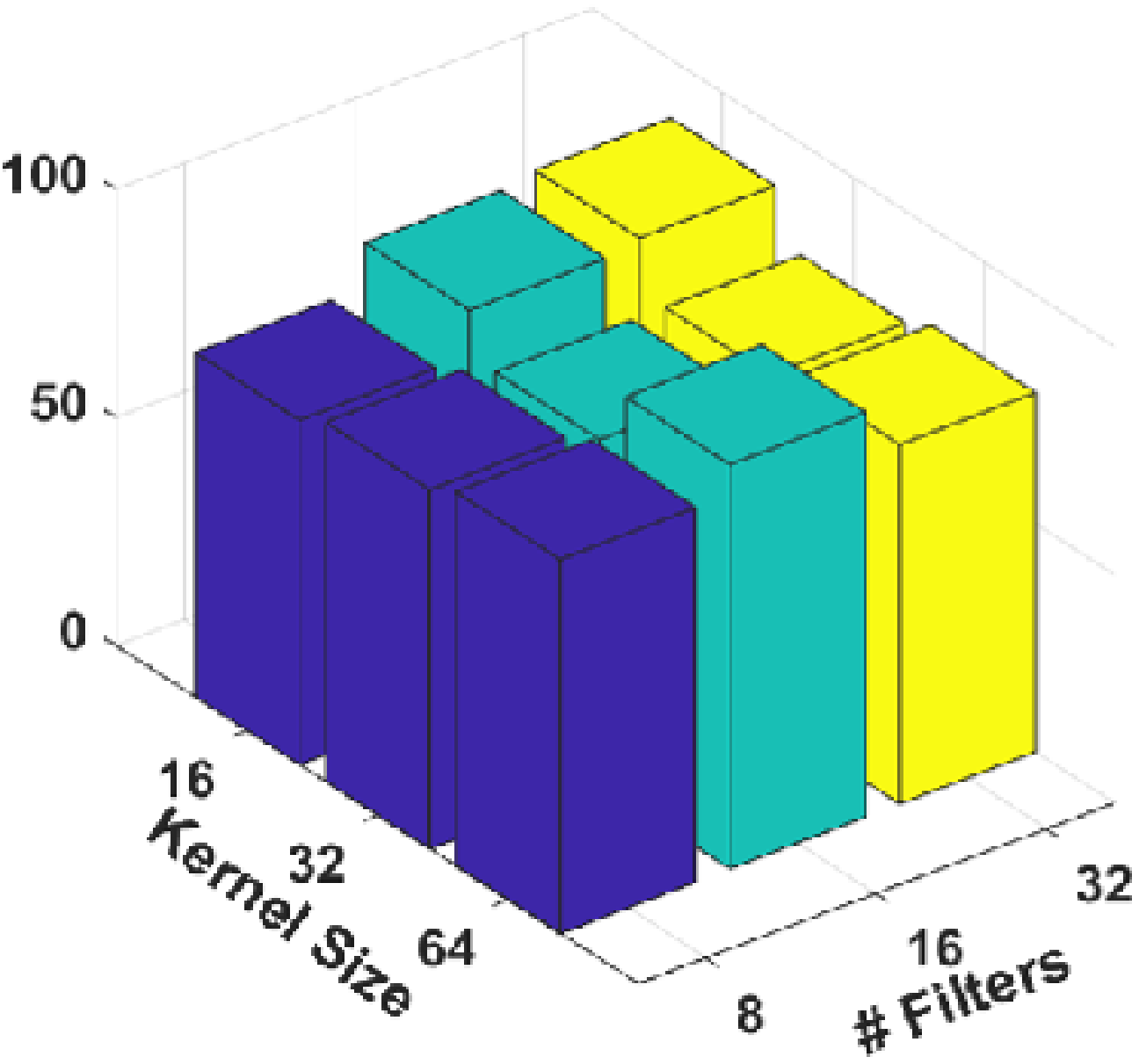}%
        \label{fig:par_mi_cat:hd16}%
    }
    \subfigure[32 hidden nodes]{
        \includegraphics[width=0.15\textwidth]{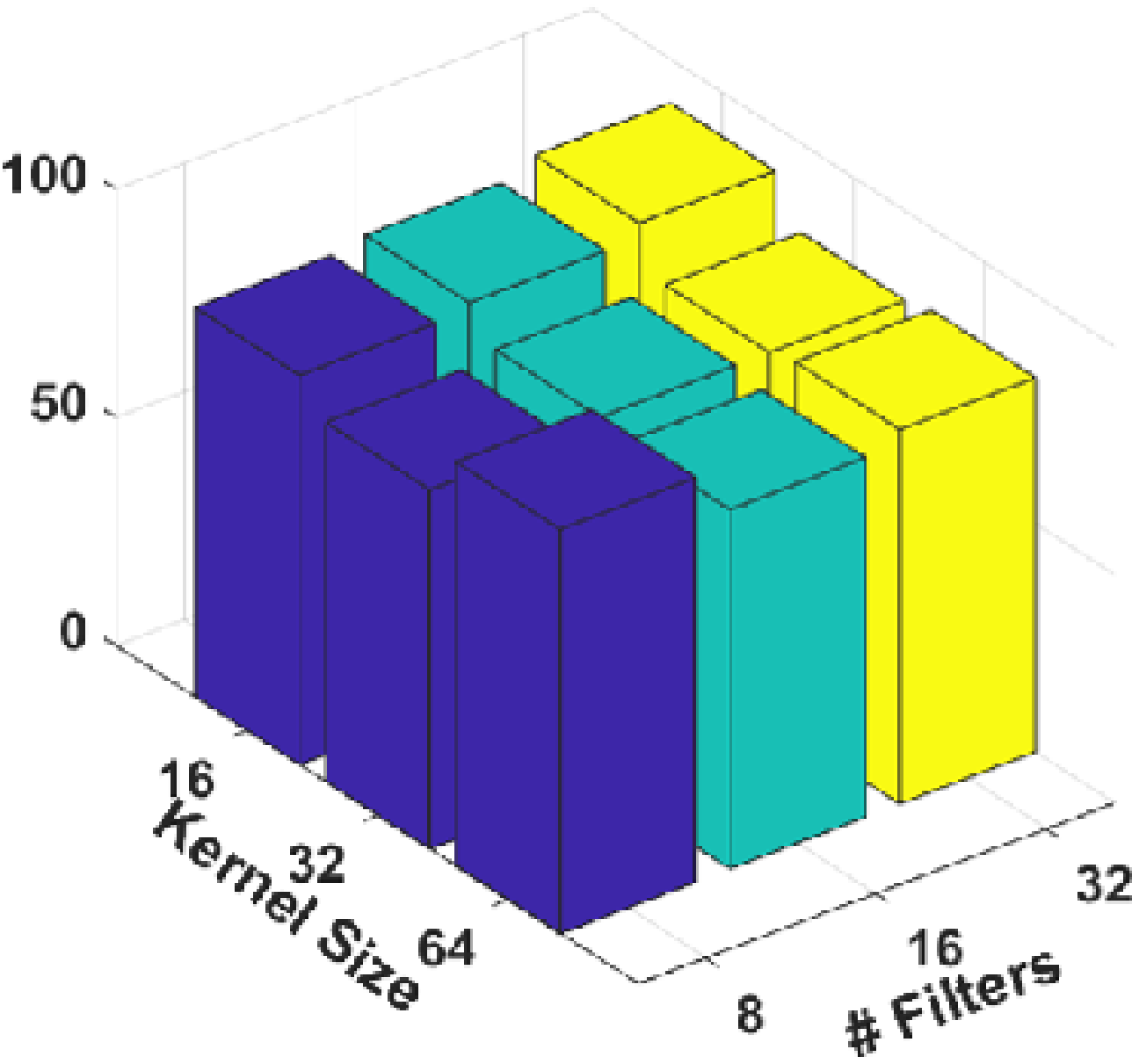}%
        \label{fig:par_mi_cat:hd32}
    }
    \vskip -1.5em
    \caption{Parameter analysis of  LAXCAT  performed on the Movement data set.} 
    \label{fig:par_mi_cat}
    \vspace{-10pt}
\end{figure}

\subsection{Ablation Study}
We also conducted an ablation study to examine the relative contributions of variable attention and temporal attention in LAXCAT. Specifically, we remove the variable attention module and obtain local context embeddings by averaging the feature vectors in each time interval (the second row in Table~\ref{tbl:ablation}). Similarly, we remove the temporal attention module and obtain summary embedding vector by averaging over the context embedding vectors (the third row in Table~\ref{tbl:ablation}). Lastly, we remove both attention modules (the last row of Table~\ref{tbl:ablation}). We conclude that both variable and temporal attention modules contribute to improved classification accuracy of LAXCAT.

\begin{table}[]
    \caption{Ablation study on variable (var.) attention and temporal time-interval (temp.)  attention. The performance of LAXCAT without variable attention is presented in row 2, that without temporal attention in row 3, and that without both attentions in the last row.}
    \label{tbl:ablation}
    \vskip -1.5em
    \begin{tabular}{r|cc} \hline
                             & \multicolumn{1}{c}{PM2.5w/} & \multicolumn{1}{c}{PM2.5w/o} \\ \toprule
    LAXCAT                   & 50.66                       & 45.53                             \\
    LAXCAT - var. attention  & 40.35                       & 42.50                              \\
    LAXCAT - temp. attention & 41.67                       & 41.71                          \\
    LAXCAT - both attentions & 40.26                       & 38.68                              \\ \bottomrule
    \end{tabular}
    \vspace{-10pt}
\end{table}

\subsection{Parameter Sensitivity Analysis}
We report results of a parameter analysis of LAXCAT. Specifically, we investigated how the kernel size (interval length), number of filters, and number of hidden nodes in the classifier neural network affect classification accuracy. Due to space limitation, we only report the results on the Movement data set, shown in Figure~\ref{fig:par_mi_cat}. The results show no clear pattern as to how the numbers of filters and hidden nodes affect the predictive performance. As for kernel size, 16 and 64 consistently yield better results than 32. When we set the kernel size to 1 (corresponding to time point based temporal attention) while fixing number of filters and hidden nodes to 8, the classification accuracy falls to around 75\%, which further underscores  the benefits of interval-based temporal attention.

\section{Summary}
We considered the problem of multivariate time series classification, in settings where in addition to achieving high accuracy, it is important to identify not only the key variables that drive the classification, but also the time intervals during which their values provide information that helps discriminate between the classes. We introduced LAXCAT, a novel, modular architecture for explainable multivariate time series classification. LAXCAT consists of a convolution-based feature extraction along with a variable based and a temporal interval based attention mechanism. LAXCAT is trained to optimize an objective function that  optimizes classification accuracy while simultaneously selecting variables as well as time intervals over which the pattern of values they assume drive the classifier output.  We present results of extensive experiments with several benchmark data sets that the proposed method outperforms the state-of-the-art baseline methods for multi-variate time series classification. The results of our case studies demonstrate that the variables and time intervals that the method identifies make sense relative to the available  domain knowledge. Some directions for ongoing and future research include generalizations of the LAXCAT framework to the settings with transfer learning~\cite{zhang2019metapred, tang2020transferring}, multi-modal~\cite{zhou2019improving, ekambaram2020attention} or multi-view~\cite{yuan2018muvan, sun2018multi}, sparsely and irregularly observed, multi-scale, MTS data~\cite{fitzmaurice2012applied, liang2020lmlfm}.

\subsection*{{\bf Acknowledgements}} 
This work was funded in part by the  NIH NCATS  grant UL1 TR002014 and by NSF  grants 2041759, 1640834, and 1636795, the Edward Frymoyer Endowed Professorship  at Pennsylvania State University and the Sudha Murty Distinguished Visiting Chair in Neurocomputing and Data Science funded by the Pratiksha Trust at the Indian Institute of Science (both held by Vasant Honavar). 
\bibliographystyle{ACM-Reference-Format}
\bibliography{ref.bib,bibliography.bib}


\end{document}